\pdfoutput=1
\documentclass[conference]{IEEEtran}

\usepackage{cite}
\usepackage{csquotes}
\usepackage{float}
\usepackage{amsmath,amssymb,amsfonts}
\usepackage{algorithmic}
\usepackage{graphicx}
\usepackage{textcomp}
\usepackage[svgnames, table, pdftex, dvipsnames]{xcolor}
\usepackage{xspace}
\usepackage{hyperref}
\usepackage{flushend}
\usepackage{tikz}
\usetikzlibrary{svg.path}

\newcommand\copyrighttext{%
  \footnotesize \textcopyright~2023 IEEE. Personal use of this material is permitted.
  Permission from IEEE must be obtained for all other uses, in any current or future
  media, including reprinting/republishing this material for advertising or promotional
  purposes, creating new collective works, for resale or redistribution to servers or
  lists, or reuse of any copyrighted component of this work in other works.
  }
\newcommand\copyrightnotice{%
\begin{tikzpicture}[remember picture,overlay]
\node[anchor=south,yshift=20pt] at (current page.south) {\fbox{\parbox{\dimexpr\textwidth-\fboxsep-\fboxrule\relax}{\copyrighttext}}};
\end{tikzpicture}%
}

\begin{document}

\title{Revisiting the Performance-Explainability Trade-Off in Explainable Artificial Intelligence (XAI)}

\author{\IEEEauthorblockN{
Barnaby Crook\IEEEauthorrefmark{1},
Maximilian Schl\"{u}ter\IEEEauthorrefmark{2},
Timo Speith\IEEEauthorrefmark{1}\IEEEauthorrefmark{3}}
\IEEEauthorblockA{\IEEEauthorrefmark{1}University of Bayreuth, Department of Philosophy, Bayreuth, Germany}
\IEEEauthorblockA{\IEEEauthorrefmark{2}Technical University Dortmund, Chair for Programming Systems, Dortmund, Germany}
\IEEEauthorblockA{\IEEEauthorrefmark{3}Saarland University, Center for Perspicuous Computing, Saarbr\"{u}cken, Germany}

Email: \{barnaby.crook, timo.speith\}@uni-bayreuth.de, maximilian.schlueter@tu-dortmund.de
}

\maketitle

\copyrightnotice
\vspace{-2ex}

\begin{abstract}
Within the field of Requirements Engineering (RE), the increasing significance of Explainable Artificial Intelligence (XAI) in aligning AI-supported systems with user needs, societal expectations, and regulatory standards has garnered recognition. 
In general, explainability has emerged as an important non-functional requirement that impacts system quality. However, the supposed trade-off between explainability and performance challenges the presumed positive influence of explainability. 
If meeting the requirement of explainability entails a reduction in system performance, then careful consideration must be given to which of these quality aspects takes precedence and how to compromise between them.
In this paper, we critically examine the alleged trade-off. We argue that it is best approached in a nuanced way that incorporates resource availability, domain characteristics, and considerations of risk.
By providing a foundation for future research and best practices, this work aims to advance the field of RE for AI.
\end{abstract}


\section{Introduction}

The field of Requirements Engineering (RE) plays a pivotal role in ensuring that artificial computing systems are aligned with user needs, societal expectations, and regulatory standards. 
As Artificial Intelligence (AI) systems become increasingly prevalent in modern society, ensuring that these systems align with customer expectations is of utmost importance. 
In this regard, Explainable AI (XAI) has emerged as a crucial factor in the pursuit of building transparent and accountable AI systems \cite{BarredoArrieta2020Explainable, Baum2022Responsibility, Langer2021What}. 
In RE, the concept of explainability has received a lot of attention recently, and it is rapidly establishing itself as a vital non-functional requirement (NFR) \cite{Chazette2021Exploring, Chazette2020Explainability, Brunotte2022Explainability, Koehl2019Explainability}.

As the demand for explainability continues to grow, it is imperative for the RE community to understand this NFR in detail. 
In this paper, we aim to contribute to this understanding by critically examining the often-proclaimed trade-off between \emph{performance} and \emph{explainability} in AI and considering its implications for RE.\footnote{The name given to the trade-off varies; some speak of an \emph{accuracy-inter\-pre\-tability} trade-off (e.g., \cite{Rudin2019Stop}). We choose the term \emph{performance} over \emph{accuracy} for maximal generality. Strictly speaking, accuracy is just one of a number of performance metrics that researchers might choose to evaluate a model. Also, we will only speak of \emph{explainability} and remain agnostic as to whether there are any differences to \emph{interpretability} (see \cite{Clinciu2019Survey, Sterz2021Towards, Speith2022Review} for discussions).}
The trade-off captures the idea that AI models which achieve better performance are less explainable, potentially leading to misalignment with user needs. 

While the trade-off is often taken for granted in the literature (e.g., \cite{BarredoArrieta2020Explainable, Gunning2019XAI, Dovsilovic2018Explainable}), there are also voices that question its existence and worry that unreflective acceptance of it can lead to negative consequences (e.g., \cite{Rudin2019Stop, Gosiewska2021Simpler, Herzog2022Ethical}). 
Thus, the question arises as to whether this trade-off actually exists and, if so, how to reconcile it with user expectations.

This article aims to explore the performance-explainability trade-off in the context of AI from the perspective of RE. 
Contrary to polarized viewpoints which either assert the trade-off as a simple fact or deny its existence altogether, we argue for a more nuanced picture that incorporates development constraints, variations in domain complexity, and an accurate appraisal of the potential of XAI. This allows us to clarify the relationship between performance and explainability. 
Overall, our aim is to provide a foundation for future research and best practices in RE for AI.

We proceed as follows. In \autoref{sec:tradeoff}, we will describe and contextualize the performance-explainability trade-off more fully. 
First, we introduce the rationale for the trade-off, before turning to arguments presented against it in an influential paper by Cynthia Rudin \cite{Rudin2019Stop}. 
Having briefly introduced her argument, \autoref{sec:conrud} is devoted to a critique of it. 
By examining several aspects of her argument in detail, we show that, while her criticism contains important insights, the view of the relationship between performance and explainability she ultimately presents is incomplete and misleading.
We end the article in \autoref{sec:discussion} by presenting our positive vision for how the RE community should think about the trade-off.

\section{The Performance-Explainability Trade-Off}
\label{sec:tradeoff}

While the performance-explainability trade-off is often presented as a fact, the reality is more complicated \cite{Bell2022It}. In this section, we describe the trade-off in more detail and summarize Rudin's critique of it.

\subsection{The Rationale for the Trade-Off}

In general, large and complex models, such as Deep Neural Networks (DNNs) or ensemble methods, often achieve high scores on performance metrics such as raw predictive accuracy, F1 score, or (negative) log loss. 
This pattern can be observed in many ML breakthroughs of the past decades (e.g., in computer vision \cite{Lawrence1997Face, LeCun1998gradient, Farabet2013Learning, Krizhevsky2017Imagenet, He2016Deep}, speech recognition \cite{Hinton2012Deep}, natural language processing \cite{Sutskever2014Sequence, Vaswani2017Attention, Bubeck2023Sparks}, and reinforcement learning \cite{Silver2016Mastering, Vinyals2019Grandmaster}). 

However, these models tend to suffer from a lack of explainability, as it is difficult to understand how they arrive at their decisions \cite{Sullivan2022Understanding}. 
Since models like DNNs have many parameters with values learned through an automated training procedure, they function as \enquote{black boxes}, where their decision-making processes are not easily expressible in human terms \cite{Zednik2021Solving, Guidotti2019Survey}.

On the other hand, models that are inherently explainable (i.e., \emph{ante-hoc explainable} models \cite{Speith2022Review}), such as rule-based systems, decision trees or linear classifiers, are simpler and more transparent.
Generally speaking, they employ explicit rules or engineered features for their decisions, making it easier for humans to understand the model's reasoning process. 

There are several factors which contribute to the explainability of any particular model \cite{Mann2023Sources}. One obvious factor is \emph{size}, with larger models often being assumed to be more difficult to understand \cite{Freitas2014Comprehensible}. 
Another factor to consider is the \emph{type of representations} a model uses \cite{Huysmans2011Empirical}. Certain representations, such as the nodes of decision trees, tend to be more understandable compared to the distributed representations learned by DNNs \cite{Samek2019Explainable}. 
A final, related factor is the complexity of the \emph{interaction} between the features of a model. 
It is easier to understand the contribution of features in a linear classifier, for instance, compared to comprehending how features in a DNN, which may interact non-linearly, influence the model's behavior \cite{Mittelstadt2019Explaining}.

Thus, based on the model that one chooses, it seems like one has to trade-off achieving the best possible performance against having a model that is fully explainable \cite{Gunning2019XAI, Dovsilovic2018Explainable, BarredoArrieta2020Explainable}.

Although the existence of a performance-explainability trade-off \emph{in theory} (e.g., for artificial data sets) is interesting, the relevant question for RE is whether we must trade these qualities off against one another in real-world contexts. Given this crucial proviso, we can express the claim of the Performance-Explainability Trade-off (PET) as follows:

\begin{description}
\item[\hspace{-2ex}(PET)] For some real-world tasks, when developing a model to perform (part of) the task, increasing the performance of the model necessarily decreases the explainability of the model (and vice versa).
\end{description}

\subsection{Rudin’s Critique of the Trade-off}

Rudin argues that, when it comes to high-stakes decisions, we are better off developing inherently explainable AI systems rather than using black-box models and explaining them after they are trained (i.e., \emph{post-hoc explainability} \cite{Speith2022Review}) \cite{Rudin2019Stop}. 

She provides several interconnected reasons for adopting this position. Initially, Rudin agrees with proponents of the PET on the premise that black-box models like DNNs are not (ante-hoc) explainable. However, unlike XAI researchers, she takes a dim view of the prospects of explaining these models with post-hoc explainability techniques, claiming that the explanatory information these techniques produce is 1) not faithful to the original models, 2) incomplete, and 3) unhelpful in most contexts. Furthermore, Rudin thinks the performance of simpler models can be improved using techniques like knowledge discovery \cite{Maimon2005Data} and feature engineering \cite{Liu1998Feature}.

Crucially, according to Rudin, the trend towards using such opaque models arises from an unwarranted belief in the PET.
Against the PET, Rudin argues that explainable models can be as good as or even outperform black-box models in most real-world applications. 
While some authors argue that the use of complex models is justified because they find \enquote{hidden patterns} in data \cite{Ha2021Unraveling, Bengio2013Representation}, Rudin argues that any patterns required for high performance could also be exploited by an ante-hoc explainable model. 

In the most general statement of her argument, Rudin claims that, for any given task, the set of almost-equally-performant models typically includes at least one simple and explainable model. This so-called \emph{Rashomon Set} argument supports Rudin’s overall proposal, which is that we should spend more time researching and engineering inherently explainable models. In Rudin’s view, this will not lead to lower performance, and thus avoids the trade-off.

Our aim is not to directly disagree with Rudin’s conclusion, but rather to address some issues in the reasoning about the PET she employs to support her claims. In particular, we will critically discuss Rudin’s stance on three topics which contribute to a nuanced appraisal of the PET: A) the potential of post-hoc explainability methods, B) the limits of iterative feature engineering, and C) the scope and plausibility of the Rashomon Set argument.

\section{Contra Rudin's Critique}
\label{sec:conrud}
We will now critically analyze Rudin's argument. To do so, we will examine each of her subclaims individually.

\subsection{The Potential of Post-Hoc Explainability Approaches}

First, we want to examine Rudin's assessment of the potential of post-hoc explainability approaches. 
We think that she underestimates this potential in terms of three properties, namely, the approaches' \emph{fidelity}, \emph{completeness}, and \emph{usefulness}.

\paragraph{Fidelity}

Let us first discuss Rudin's take on the \emph{fidelity} of post-hoc explainability approaches. Following her discussion, we assume that fidelity means that an explanation accurately captures the decision processes of a model \cite{Speith2022Evaluate}; i.e., that it reflects how and what the model computes \cite{Molnar2019Interpretable}. 

In her argument, Rudin asserts that explaining an AI model after it has been trained (i.e., post-hoc explainability) must result in explanations that lack fidelity to the original model. 
Further, she argues that these explanations should not really be called \enquote{explanations} at all, since they could respond to completely different features than the original model.
According to her, this is not morally acceptable, especially in situations where consequential human values are at stake. 
For example, a person might be denied medical treatment due to a racial bias in the model, but this could be (un)intentionally obscured by explanations that are not fidelitous to the model.

While Rudin's concerns may hold true for \emph{model-agnostic} explainability approaches (i.e., approaches that explain a model without regards to its internals or structure, e.g., by mere input/output analysis \cite{Speith2022Review}), it fails to consider the potential of \emph{model-specific} ones (i.e., approaches that explain a model with at least some recourse to the model's internals or structure \cite{Speith2022Review}) to provide fidelitous explanations. 

It is true that there is a lack of research that addresses the fidelity of information generated by explainability approaches. 
This research gap can be attributed, in part, to the absence of formal and technical criteria for assessing fidelity \cite{Molnar2019Interpretable, Adebayo2018Sanity, Amparore2021Trust}. 

Recently, however, researchers have started to address this issue \cite{Amorim2023Evaluating}. Adebayo et al. \cite{Adebayo2018Sanity}, for instance, proposed a sanity check specifically designed for approaches that generate saliency maps. Such checks can be used to assess whether explainability approaches are sensitive to class-relevant features, thus providing evidence against the worry that they lack fidelity.
By applying the check, Abedayo et al. assessed various approaches and identified variations in fidelity. Notably, while some model-specific approaches such as \emph{Vanilla Backpropagation} \cite{Simonyan2013Deep} and \emph{Grad-CAM} \cite{Selvaraju2017Grad-CAM} demonstrated fidelity by passing the check, others like \emph{Guided Backpropagation} \cite{Springenberg2015Striving} did not.

In light of these findings, it becomes evident that Rudin's claim requires a more nuanced perspective. 
By evaluating the fidelity of explanations it is possible to distinguish between explainability approaches that offer accurate representations of decision-making processes and those that do not.  
Further exploration and evaluation of model-specific explainability approaches are necessary to gain a deeper understanding of their fidelity and potential to provide reliable insights into AI models.

\paragraph{Completeness}
Admittedly, Rudin's criticism may not only concern the fidelity of explanations, but also their \emph{completeness} (see \cite{Zhou2021Evaluating} for a discussion): that \emph{all} factors playing into a model's output are cited.

If we analyze the completeness of saliency maps, the situation does not look as good as it did with their fidelity \cite{Dombrowski2019Explanations}. This is because saliency maps are mathematically capable of responding only to the most important features, and it cannot be ruled out that some comparably less important features also had an impact on the output of a model. However, such features could be crucial, for example, when it comes to assessing possible biases regarding protected attributes \cite{Tschantz2022Proxy}. 

Saliency maps, however, are not the only type of post-hoc explainability approach that is potentially fidelitous. In particular, there are several approaches that attempt to exploit the latent information contained in a DNN for explanations, such as feature visualization \cite{Olah2017Feature, Olah2018Building} and Testing with Concept Activation Vectors (TCAV) \cite{Kim2018Interpretability}. Feature visualization, for instance, aims to make the role of neurons or layers in DNNs clear by synthesizing images that maximize their activation \cite{Olah2017Feature, Nguyen2019Understanding}. Through systematic use of these techniques, researchers are able to build up fidelitous and complete explanations of the internal structure of opaque models \cite{Cammarata2020Thread}. 

Furthermore, incomplete explanations that are fidelitous can be sufficient in many cases, such as when the tracked features suffice to expose a model's deficiency. In these cases, incomplete explanations are even preferable to complete explanations, since the latter can overwhelm a recipient and in some cases even obscure the malfunction of a model by their sheer volume \cite{Ananny2018Seeing}. Indeed, psychological work on causal explanation has shown that recipients prefer explanations involving a limited but relevant set of causes \cite{Miller2019Explanation}.

\paragraph{Usefulness} In addition to these two critiques of post-hoc explainability approaches, Rudin also questions their usefulness. In particular, she argues that post-hoc explainability techniques that produce heatmaps (e.g., Grad-CAM) are not very useful because they do not allow for good assessments of predictions. She illustrates this with an example: the most probable class (e.g., wolf) is often hardly distinguishable from that of a totally unrelated, less probable class (e.g., flute).

While this may preclude the suitability of such explainability approaches for some applications, these heatmaps can at least suffice to detect if the model uses completely irrelevant features for the classification (see \cite{Lapuschkin2019Unmasking} for an example).

Furthermore, saliency maps are just one of a multitude of post-hoc explainability approaches that may be useful for different purposes. In addition to feature visualization, we also mentioned TCAV above. TCAV can be used to query the influence of user-defined concepts (e.g., striped, female) on a classification, allowing for the detection of arbitrary biases in DNNs. Further, counterfactual explanations can be used to find out how to achieve a favorable decision from a black-box model \cite{Karimi2021Algorithmic, Karimi2022Survey}. In summary, we think that there are post-hoc approaches that can be usefully employed to explain the output of black-box models even in high-stakes situations.

\subsection{The Limits of Iterative Feature Engineering}

One of Rudin's main claims is that, in most cases, there are ante-hoc explainable models that can achieve performance equivalent to black-box models. 
She suggests that through an iterative process of knowledge discovery and feature engineering, one can almost always arrive at such models.
However, we believe that two obstacles stand in the way of such a process: \emph{hidden patterns} and, relatedly, \emph{unengineerable features}.

\paragraph{Hidden Patterns} 
In a short but suggestive passage, Rudin proposes that one of the reasons for the widespread acceptance of the PET is a belief that opaque models exploit so-called \enquote{hidden patterns} in the data to achieve good performance. While Rudin does not explicitly define the term, we interpret it to refer to complex, non-linear, and/or distributed combinations of input features that capture predictively relevant information in the input data (see \cite{Bengio2013Representation} for a discussion).

However, Rudin expresses skepticism about the notion that black-box models, such as DNNs, can uncover hidden patterns that cannot also be used by simpler models. She thinks reliance on these opaque models can be avoided given sufficient researcher ability in crafting the right kind of model.

Rudin's discussion of hidden patterns is closely related to her other worries about black-box models. In particular, she argues that the assumption of hidden patterns is usually misleading and can give rise to false beliefs about the superior capabilities of opaque models. She believes that, often, instead of uncovering hidden patterns that reflect the genuine structure of the data domain in question, these models exploit spurious correlations in the training data, which may not necessarily reflect causality or any meaningful patterns in the real world.

Rudin raises valid concerns regarding the susceptibility of complex ML models to overfitting the training data, including noise and irrelevant features, resulting in suboptimal generalization and unreliable predictions on new data (see \cite{Belkin2019Reconciling, Yang2020Rethinking} for discussions). However, recent theoretical work suggests that neural networks often do not overfit, even when their parameters outnumber the training samples \cite{zhang2021understanding}. 
 
More fundamentally, we think it is highly likely that hidden patterns do exist in some domains. Indeed, paying close attention to how and why Deep Learning (DL) became the dominant technique in ML gives us evidence for this.

In the early stages of ML, limited computational power and small datasets restricted the complexity of ML models it was possible to train (in terms of, e.g., the number of trainable parameters) \cite{Goodfellow2016Deep}.
Thus, most solutions relied on either manually crafted features or the application of general purpose algorithms, such as (K)PCA, LDA, Fourier and Wavelet Transformation, SOM, KL Transform, Histograms, Clustering, and Embeddings, to automatically derive simple features \cite{Jain2000Statistical, Webb2002Statistical}.
However, in highly complex domains, such as image recognition and natural language processing, both of these approaches failed to achieve adequate performance \cite{LeCun2015Deep}.

By efficiently leveraging computational resources to autonomously discover their own feature spaces during training, DL techniques have drastically reduced the need for human intervention in the feature engineering process \cite{Schmidhuber2015Deep, LeCun2015Deep, Goodfellow2016Deep}. Given enough computational resources, the data-driven approaches of highly parameterized ML models significantly outperform human engineering in various ML tasks \cite{sutton2019bitter, chowdhery2022palm}.

For one prominent example, consider the recent success of large language models. Increasing the parameter count of these models led to them learning the structure of natural language well enough to produce meaningful and coherent text \cite{Brown2020Language, chowdhery2022palm}. We suggest that the most plausible explanation of this phenomenon is that increasing the scale of the models enabled them to learn subtle, complex dependencies that are difficult to articulate simply but are necessary for good performance on the difficult task of next-word prediction \cite{Kaplan2020Scaling}.

Feature spaces found by DL satisfy our criteria for hidden patterns: they are complex, non-linear, and use distributed representation.
Further, they capture information that is likely required to solve the tasks at hand, as demonstrated by their performance advantage over (ante-hoc) explainable models.

According to Rudin, even if there were hidden patterns, ante-hoc explainable models could also exploit them to achieve approximately equivalent performance. However, given these historical developments, this seems unlikely. The absence of satisfying results despite decades of research suggests simpler ML approaches may be incapable of learning the hidden patterns required for good performance in certain domains.

In sum, we think Rudin is right to consider that the question of when complex, opaque models are required is closely related to the question of which domains have \enquote{hidden patterns}. Indeed, we agree with Rudin's view that the assumption of hidden patterns and the mindless deployment of complex models can be harmful. However, we think her hope that there may be \emph{no} domains containing hidden patterns which are too complex to be learned by simple models is unlikely to be true.

\paragraph{Unengineerable Features}

As outlined above, the transition from handcrafted to learned feature spaces has enabled improved performance on more complex and challenging problems (see also \cite{LeCun2015Deep}). 
The primary advantage of DL lies in its ability to perform well on tasks where it is difficult to encode human intuition using traditional algorithms or features. 
Consequently, DL has gained significant recognition in fields such as computer vision (e.g., AlexNet \cite{Krizhevsky2017Imagenet}, ResNet \cite{He2016Deep}), natural language processing (e.g., Transformers \cite{Vaswani2017Attention}), and strategic game playing (e.g., AlphaGo \cite{silver2017mastering}, AlphaStar \cite{Vinyals2019Grandmaster}), where traditional algorithms tend to underperform.

Interestingly, DL outperforms handcrafted feature spaces even in domains where we already have a deep intuitive understanding of the problem (e.g., natural language processing).
Many of the aforementioned tasks are so essential for humankind that almost every human has the innate ability to learn to perform reasonably well given some initial training.

Nevertheless, encoding human knowledge and intuition into computer-readable formats, which usually requires a rigorous mathematical or logical representation, has proven difficult.
This may be because features have to be built for detailed, low-level inputs like pixels in computer vision or tokens in natural language processing. Unlike such inputs, human reasoning happens at a higher-level \cite{Marcus2003Algebraic}.
Finally, even if \emph{some} individual features could be extracted through a process of iterative feature engineering, the sheer quantity of such features may be too high to efficiently encode all of them by hand. 
This leads to a critical limitation in knowledge-discovery based approaches: even though we know what might constitute good features for a given ML task, we are sometimes unable to express them to computer systems.

\subsection{Rashomon Set Argument}

The Rashomon set argument \cite{Rudin2019Stop, Semenova2022Existence} proposes that simple (and thus explainable) models exist within the set of reasonably performant models (the Rashomon set) for almost every domain. The argument suggests that the reason simple models often exist is that the Rashomon set is large, containing many performant models. If there are many such models, then we should expect at least one of them to be ante-hoc explainable. 

By studying the Rashomon ratio, which measures the volume of performant models relative to the hypothesis space, one can estimate if a simple model is likely to exist before explicitly searching for it \cite{Semenova2022Existence}. If many real-world datasets have large Rashomon sets, it implies that explainable models can, in principle, be used for important decisions without sacrificing performance. In essence, the Rashomon set argument provides a framework for understanding when and why performant \emph{and} ante-hoc explainable models are likely to be found \cite{Semenova2022Existence}.

The Rashomon set argument is certainly intriguing. However, despite being presented as speaking for the existence of ante-hoc explainable models across domains, the argument is weaker than it sounds. We give three reasons why:

\paragraph{Plausibility} The conclusion that Rudin draws depends upon two critical assumption: i) that Rashomon sets will typically be large, and ii) that large Rashomon sets will contain simple models. However, neither of these assumptions are justified with rigorous argument. Accordingly, one can also reverse the claims: it may well be that the Rashomon set is not large, and, irrespective of this, it may be that the Rashomon set does not contain simple models. Although her analysis is interesting, Rudin has given us no independent reason to believe that explainable models will be contained in the Rashomon set in every domain.

\paragraph{Scope} We also note that Rudin conceives ante-hoc explainable models to refer to certain classes of models (e.g., decision trees). These models are deemed explainable because their features are usually hand-crafted and have a straightforward semantic interpretation. However, although these models may \emph{typically} be explainable, that does not mean that they will remain explainable if they are allowed to grow arbitrarily large and represent numerous fine-grained distinctions \cite{Lipton2018Mythos}. 

\paragraph{Time} Finally, even if the Rashomon set argument were valid, it would still not mean that finding ante-hoc explainable models would be easy. On the contrary, an existence proof does not mean that one will be able to find the model efficiently or even at all. While Rudin states that one could enumerate all performant models and thus eventually get to the explainable one, even finding optimal decision trees is an NP hard problem \cite{murthy1998automatic}, making it not guaranteed to get to a solution in a remotely reasonable time.

\section{Re-Imagining the Performance-Explainability Trade-Off}
\label{sec:discussion}

Overall, though we find the thrust of Rudin’s criticisms of the PET insightful, some of her claims are inadequately supported, unlikely to apply across domains, or make assumptions biased towards the conclusion she favors. In light of this, we sketch a more nuanced view that incorporates the valuable aspects of Rudin’s criticisms but also accounts for its weaknesses. We start with expanding the PET.

\subsection{From PET to PET+: Incorporating Further Factors} 

The discussion of the Rashomon set argument revealed that, even if it were true in theory, one would be required to expend a great deal of time enumerating and evaluating models in order to obtain an explainable and performant one. This is not the only point in Rudin's argument where development resources come into play. Her proposal of an iterative feature engineering process also demands a significant time investment as well as researcher expertise and supporting resources (e.g., tools, money, etc.). 

However, when Rudin considers the use of black-box models explained with post-hoc techniques, she does not account for variation in this additional dimension. In particular, she does not consider how well black-box models could be explained given development resources comparable to those required for iterative feature engineering (e.g., by systematically reverse engineering complex models with model-specific explainability techniques). To compare black-box methods and iterative feature engineering fairly, however, we need to consider what these two approaches can achieve given \emph{the same} amount of resources. We think this is a crucial point and suggest that the PET needs to be refined to accommodate the availability of resources.

\begin{figure}[H]
    \centering
    \vspace{-1ex}
    \includegraphics[width=0.32\textwidth]{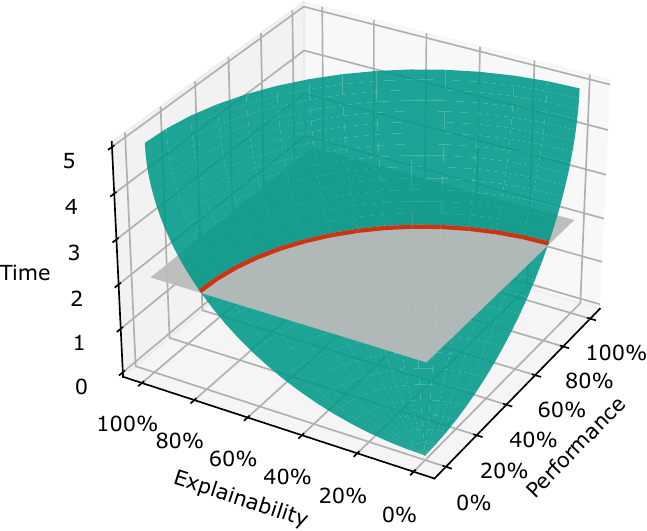}
    \caption{Idealized depiction of PET+ as a trade-off with the three dimensions  \emph{explainability}, \emph{performance}, and \emph{time}. With more time available, better solutions can be found. For a given temporal budget (e.g., grey plane), however, a choice has to be made between the solutions of the Pareto set (red curve).}
    \label{fig:pareto_front}
    \vspace{-1ex}
\end{figure}

Thus, we suggest that the availability of developmental resources, especially time, constitutes an extra dimension for the PET. Making the simplifying assumption that expertise and financial budget are held fixed, we attain a three-factor trade-off of time, performance, and explainability: PET+.

\begin{description}
\item[\hspace{-2ex}(PET+)] For some real-world tasks, when developing a model to perform (part of) the task, increasing the performance of the model necessarily decreases the explainability of the model (and vice versa) given the same temporal resources. By increasing the temporal budget, we may achieve a Pareto improvement \cite{Ehrgott2005Multicriteria} on our previous best model (see \autoref{fig:pareto_front}).
\end{description}

\begin{figure*}[ht]
    \centering
    \includegraphics[width=0.9\textwidth]{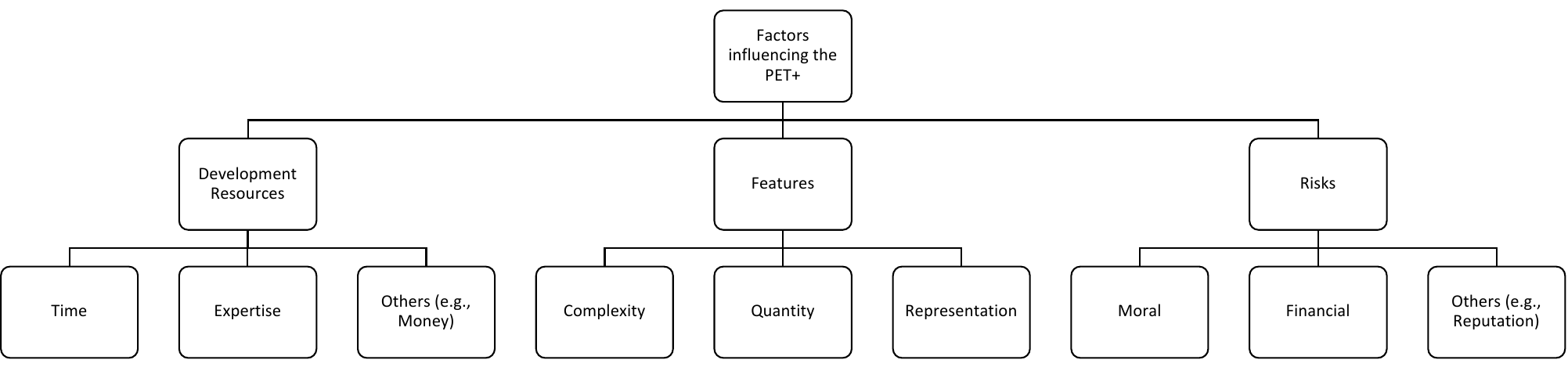}
    \caption{Our classification of the various factors that influence the PET+. The levels of performance and explainability it is possible to achieve are constrained by the development resources (e.g., time, expertise) available and the properties of the features (i.e., their complexity, quantity, and representation) that exist in the particular domain. Risks (e.g., moral, financial) affect the relative importance of performance and explainability in any given real-world scenario.}
    \label{fig:my_label}
    \vspace{-1ex}
\end{figure*}

We suggest that, in many cases, developing explainable models that are also highly performant requires extensive development resources like time, suitably experienced researchers, and special tools. In general, we agree with Rudin's claim that high-stakes domains justify the deployment of such resources. As such, we think PET+ is a more appropriate characterization of the issue that (requirements) engineers are facing than PET. However, we think this holds true for \emph{both} Rudin's preferred method of designing ante-hoc explainable models \emph{and} approaches which rely on training black-box models and then explaining them afterwards. Thus, it is crucial to apply this consideration equally to \emph{all} possible approaches to engineering ML models. Only by doing this can the strengths and weaknesses of different approaches be accurately compared.

Finally, we note that the three-factor trade-off captured by PET+ still does not describe the full complexity of model development. Due to Rudin's focus on high-stake situations, she discourages the use of explanations lacking fidelity and appeals for the use of explainable models partially for \emph{moral} reasons. We agree with Rudin that moral considerations should come into play when deciding which values to prioritize in model-building. Indeed, we think that moral risks should be considered alongside other risks that developers take on such as financial risks and risks to one's reputation.

\subsection{PET+ as a Multi-Objective Optimization Problem}

Overall, the performance-explainability trade-off is influenced by many factors (e.g., development resources, domain details, and risks; see \autoref{fig:my_label}). Further, which outcomes are considered most desirable in each real-world scenario depends critically on the relative importance we assign to performance and explainability. To decide how to proceed in light of this, it will be beneficial to have a framework that incorporates all of the relevant factors. To this end, we suggest to cast PET+ as a multi-objective optimization problem as follows.

Consider a specific decision (e.g., making medical diagnoses) which, at $t=0$, is made by a system (which may be a human) that produces some cost at every time-step. The cost is inversely proportional to a linear combination of performance and explainability (i.e., the better the performance and the greater the explainability of the model, the lower the cost). 

With this setup in mind, researchers aim to reduce the cost by evaluating and selecting a sequence of actions over time (i.e., developing, refining, explaining, and ultimately, deploying new models). To choose which actions to take, they need to estimate the value (i.e., reduced cost relative to some baseline) of different sequences of actions based on their expertise and domain knowledge. This value will be proportional to the improvements to performance and explainability they can produce, as well as how quickly they can produce those improvements. The researchers' ultimate goal is to minimize the \emph{level of regret}, defined as the difference between the accumulated cost-over-time and the cost that would be incurred by the best sequence of actions. 

Each of the factors we have discussed is captured by this formulation of the problem. First, consider \emph{development resources}. The specific \emph{expertise} researchers bring to a problem constrains what level of performance and explainability they might achieve. In our framework, researcher skills can be thought of as restrictions on the set of actions available. Similarly, researcher knowledge (e.g., about the nature of the relationship between symptoms and various medical conditions) constrains how accurately the value of different sequences of actions can be calculated. At the same time, temporal constraints limit the length of action sequences that are possible, and, thereby, the level of performance and explainability it is possible to achieve.

Next, consider what kind of \emph{features} are present in the domain. If the features which must be learned to achieve good performance are complex, numerous, and best represented in a distributed way (i.e., there are \emph{hidden patterns}), it may be that constructing black-box models and then explaining them post-hoc is the best approach. Where they are absent, an iterative feature engineering approach may be preferable. In our framework, these feature-based considerations constrain the true value of different actions (i.e., how much they actually improve performance and explainability). 

Finally, \emph{risk} enters the framework in the relative weighting of performance and explainability. Depending on which values are deemed to be most important in the real-world scenario in question, the weights will be set differently, thus altering how the cost is calculated. For example, explainability may be particularly important in a judicial setting \cite{eliot2021need}, while performance might be prioritized in a medical scenario \cite{London2019Artificial}. 

Practically, we think that model development, in light of the PET+, can be usefully informed by asking the following heuristic questions: What expertise and tools are available? What is the temporal and financial budget? What properties does the data domain in question have? How important are performance and explainability relative to one another? Given a set of answers to these questions, one should consider which actions to take to minimize expected regret.

\section{Conclusion}

Our reflections on the performance-explainability trade-off are intended to serve as a catalyst for further discussion and exploration within the RE community. We believe that by painting a more nuanced picture of the trade-off, captured by PET+, we can inspire researchers, practitioners, and stakeholders to collaborate in shaping the future of RE for AI systems.

\section*{Acknowledgments}
Work on this paper was partially funded by the Volkswagen Foundation grants \textsc{AZ 98509} and \textsc{AZ 98514} \href{https://explainable-intelligent.systems}{\enquote{Explainable Intelligent Systems}} (\textsc{EIS}) and by the \textsc{DFG} grant 389792660 as part of \href{https://perspicuous-computing.science}{\textsc{TRR}~248}. We thank the EIS reading group and three anonymous reviewers for their helpful feedback.

\bibliographystyle{IEEEtran}
\bibliography{bibliography}

\end{document}